\documentclass[11pt]{article}

\usepackage{authblk}
\usepackage{konvens2018}

\usepackage[hyphens]{url}
\usepackage{fixltx2e}

\usepackage{amsmath,amsfonts,amssymb}
\usepackage{algorithm}
\usepackage[noend]{algorithmic}

\usepackage{booktabs}
\usepackage{multirow}

\DeclareMathAlphabet{\mathcal}{OMS}{cmsy}{m}{n}

\newcommand{\REQUIREP}{\item[\hphantom{\textbf{Input:}}]}

\newcommand{\watset}{\textsc{Watset}}

\newcommand\sense[2]{\textit{#1}\ensuremath{^{#2}}}

\DeclareMathOperator{\tf}{\mathrm{tf}}
\DeclareMathOperator{\idf}{\mathrm{idf}}
\DeclareMathOperator{\tfidf}{\tf\!\textrm{--}\!\idf}

\DeclareMathOperator{\senses}{\mathrm{senses}}
\DeclareMathOperator{\words}{\mathrm{words}}

\DeclareMathOperator{\hlabel}{\mathrm{label}}
\DeclareMathOperator{\ssim}{\mathrm{sim}}
\DeclareMathOperator{\NN}{\mathrm{NN}}

\usepackage{todonotes}

\title{Unsupervised Sense-Aware Hypernymy Extraction}

\author[$\dag$]{\textbf{Dmitry Ustalov}}
\author[$\ddag$]{\textbf{Alexander Panchenko}}
\author[$\ddag$]{\\\textbf{Chris Biemann}}
\author[$\dag$]{\textbf{Simone Paolo Ponzetto}}

\affil[$\dag$]{University of Mannheim, Germany}
\affil[ ]{\texttt{\{dmitry,simone\}@informatik.uni-mannheim.de}}

\affil[$\ddag$]{University of Hamburg, Germany}
\affil[ ]{\texttt{\{panchenko,biemann\}@informatik.uni-hamburg.de}}

\date{}

\begin{document}

\maketitle

\begin{abstract}
In this paper, we show how unsupervised sense representations can be used to improve hypernymy extraction. We present a method for extracting disambiguated hypernymy relationships that propagate hypernyms to sets of synonyms (synsets), constructs embeddings for these sets, and establishes sense-aware relationships between matching synsets. Evaluation on two gold standard datasets for English and Russian shows that the method successfully recognizes hypernymy relationships that cannot be found with standard Hearst patterns and Wiktionary datasets for the respective languages.
\end{abstract}

\section{Introduction}

Hypernymy relationships are of central importance in natural language processing. They can be used to automatically construct taxonomies~\cite{Bordea:16,faralli17,faralli18}, expand search engine queries~\cite{Gong:05}, improve semantic role labeling~\cite{Shi:05}, perform generalizations of entities mentioned in questions~\cite{Zhou:13}, and so forth. One of the important use cases of hypernyms is lexical expansion as in the following sentence: ``This bar serves fresh \textit{jabuticaba} juice''. Representation of the rare word ``jabuticaba'' can be noisy, yet it can be substituted by its hypernym ``fruit'', which is frequent and has a related meaning. Note that, in this case, sub-word information provided by character-based distributional models, such as \textit{fastText}~\cite{bojanowski2017enriching}, does not help to derive the meaning of the rare word.  

Currently available hypernymy extraction methods perform extraction of hypernymy relationships from text between two ambiguous words, e.g., $\text{apple} \succ \text{fruit}$. However, by definition in \newcite{Cruse:86}, hypernymy is a binary relationship between senses, e.g., $\sense{apple}{2} \succ \sense{fruit}{1}$, where $\sense{apple}{2}$ is the ``food'' sense of the word ``apple''. In turn, the word ``apple'' can be represented by multiple lexical units, e.g., ``apple'' or ``pomiculture''. This sense is distinct from the ``company'' sense of the word ``apple'', which can be denoted as $\sense{apple}{3}$. Thus, more generally, hypernymy is a relation defined on two sets of disambiguated words; this modeling principle was also implemented in WordNet~\cite{Fellbaum:98}, where hypernymy relations link not words directly, but instead synsets. This essential property of hypernymy is however not used or modeled in the majority of current hypernymy extraction approaches. In this paper, we present an approach that addresses this shortcoming.

The contribution of our work is a novel approach that, given a database of noisy ambiguous hypernyms, (1) removes incorrect hypernyms and adds missing ones, and (2) disambiguates related words.  Our unsupervised method relies on synsets induced automatically from synonymy dictionaries. In contrast to prior approaches, such as the one by \newcite{Pennacchiotti:06}, our method not only disambiguates the hypernyms but also extracts new relationships, substantially improving F-score over the original extraction in the input collection of hypernyms. We are the first to use sense representations to improve hypernymy extraction, as opposed to prior art.

\begin{figure*}[t]
  \centering
  \includegraphics[width=.875\textwidth]{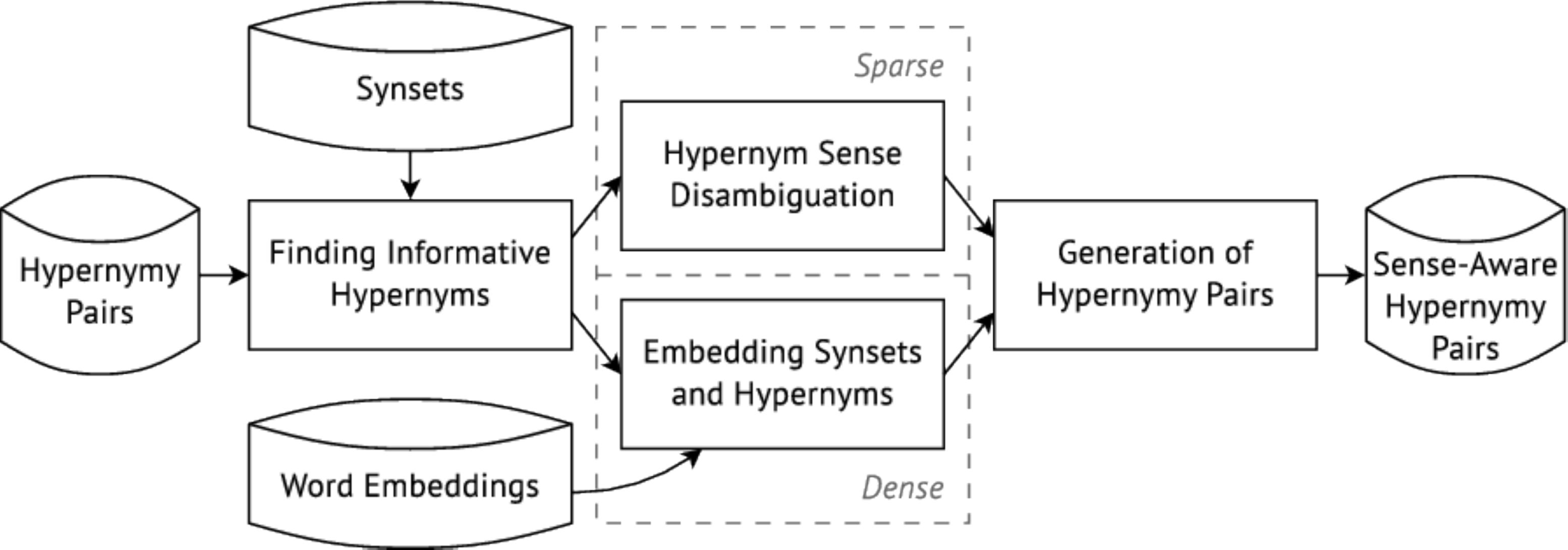} 
  \caption{\label{fig:pipeline}Outline of the proposed method for sense-aware hypernymy extraction using synsets.}
\end{figure*}

\section{Related Work}

In her pioneering work, \newcite{Hearst:92} proposed to extract hypernyms based on lexical-syntactic patterns from text. \newcite{Snow:04} learned such patterns automatically, based on a set of hyponym-hypernym pairs. \newcite{Pantel:06} presented another approach for weakly supervised extraction of similar extraction patterns. All of these approaches use a small set of training hypernymy pairs to bootstrap the pattern discovery process. \newcite{TjongKimSang:07} used Web snippets as a corpus for a similar approach. More recent approaches exploring the use of distributional word representations for extraction of hypernyms and co-hyponyms include~\cite{Roller:14,Weeds:14,Necsulescu:15,Vylomova:16}. They rely on two distributional vectors to characterize a relationship between two words, e.g., on the basis of the difference of such vectors or their concatenation.

Recent approaches to hypernym extraction went into three directions: (1) unsupervised methods based on such huge corpora as CommonCrawl\footnote{\url{https://commoncrawl.org}} to ensure extraction coverage using \newcite{Hearst:92} patterns~\cite{Seitner:16}; (2) learning patterns in a supervised way based on a combination of syntactic patterns and distributional features in the HypeNet model~\cite{Shwartz:16}; (3) transforming~\cite{Ustalov:17:eacl} or specializing~\cite{Glavas:17} word embedding models to ensure the property of asymmetry. We tested our method based on a large-scale database of hypernyms extracted in an unsupervised way using Hearst patterns. While methods, such as those by \newcite{Mirkin:06}, \newcite{Shwartz:16}, \newcite{Ustalov:17:eacl} and \newcite{Glavas:17} use distributional features for extraction of hypernyms, they do not take into account word sense representations: this is despite hypernymy being a semantic relation holding between senses. 

The only sense-aware approach we are aware of is presented by \newcite{Pennacchiotti:06}. Given a set of extracted binary semantic relationships, this approach disambiguates them with respect to the WordNet sense inventory~\cite{Fellbaum:98}. In contrast to our work, the authors do not use the synsets to improve the coverage of the extracted relationships.

Note that we propose an approach for post-processing of hypernyms based on a model of distributional semantics. Therefore, it can be applied to any collection of hypernyms, e.g., extracted using Hearst patterns, HypeNet, etc. Since our approach outputs dense vector representations for synsets, it could be useful for addressing such tasks as knowledge base completion \cite{Bordes:11}.

\section{Using Synsets for Sense-Aware Hypernymy Extraction}

\begin{algorithm*}[t]
\caption{\label{alg:linking}Unsupervised Sense-Aware Hypernymy Extraction.}
\begin{algorithmic}[1]
\REQUIRE{a vocabulary $V$, a set of word senses $\mathcal{V}$, a set of synsets $\mathcal{S}$, a set of \textit{is-a} pairs $R \subset V^2$.}
\REQUIREP{a number of top-scored hypernyms $n \in \mathbb{N}$,}
\REQUIREP{a number of nearest neighbors $k \in \mathbb{N}$,}
\REQUIREP{a maximum matched synset size $m \in \mathbb{N}$.}
\ENSURE{a set of sense-aware \textit{is-a} pairs $\mathcal{R} \subset \mathcal{V}^2$.}
\FORALL{\label{alg:linking:label:begin}$S \in \mathcal{S}$}
\STATE \label{alg:linking:label:end}$\hlabel(S) \gets \{h \in V : (w, h) \in R, w \in \words(S)\}$
\ENDFOR
\FORALL{\label{alg:linking:tfidf:begin}$S \in \mathcal{S}$}
\FORALL{$h \in \hlabel(S)$}
\STATE \label{alg:linking:tfidf:end}$\tfidf(h, S, \mathcal{S}) \gets \tf(h, S) \times \idf(h, \mathcal{S})$
\ENDFOR
\ENDFOR
\FORALL[Hypernym Sense Disambiguation]{$S \in \mathcal{S}$}
\STATE $\widehat{\hlabel}(S) \gets \emptyset$
\FORALL[Take only top-$n$ elements of $\hlabel(S)$]{$h \in \hlabel(S)$}
\STATE \label{alg:linking:sparse:wsd}$\hat{S} \gets {\arg\max}_{S' \in \mathcal{S} : \senses(h) \cap S' \neq \emptyset} \ssim(\hlabel(S), \words(S'))$
\STATE \label{alg:linking:sparse:add:begin}$\hat{h} \gets \senses(h) \cap \hat{S}$
\STATE \label{alg:linking:sparse:add:end}$\widehat{\hlabel}(S) \gets \widehat{\hlabel}(S) \cup \{\hat{h}\}$
\ENDFOR
\ENDFOR
\FORALL[Embedding Synsets and Hypernyms]{$S \in \mathcal{S}$}
\STATE \label{alg:linking:dense:svec}$\vec{S} \gets \frac{\sum_{w \in \words(S)} \vec{w}}{|S|}$
\STATE \label{alg:linking:dense:lvec}$\overrightarrow{\hlabel}(S) \gets \frac{\sum_{h \in \hlabel(S)} \tfidf(h, S, \mathcal{S}) \cdot \vec{h}}{\sum_{h \in \hlabel(S)} \tfidf(h, S, \mathcal{S})}$
\STATE \label{alg:linking:dense:wsd}$\hat{S} \gets {\arg\max}_{S' \in \NN_k(\overrightarrow{\hlabel}(S)) \cap \mathcal{S} \setminus \{S\}} \ssim(\overrightarrow{\hlabel}(S), \vec{S}')$
\IF{\label{alg:linking:dense:size:begin}$|\hat{S}| \leq m$}
\STATE $\widehat{\hlabel}(S) \gets \widehat{\hlabel}(S) \cup \hat{S}$
\ENDIF\label{alg:linking:dense:size:end}
\ENDFOR
\RETURN \label{alg:alinking:return}$\bigcup_{S \in \mathcal{S}} S \times \widehat{\hlabel}(S)$
\end{algorithmic}
\end{algorithm*}

We use the sets of synonyms (synsets) expressed in such electronic lexical databases as WordNet~\cite{Fellbaum:98} to disambiguate the words in extracted hyponym-hypernym pairs. We also use synsets to propagate the hypernymy relationships to the relevant words not covered during hypernymy extraction. Our unsupervised method, shown in \figurename~\ref{fig:pipeline}, relies on the assumption that the words in a synset have similar hypernyms. We exploit this assumption to gather all the possible hypernyms for a synset and rank them according to their importance (Section~\ref{sub:finding}). Then, we disambiguate the hypernyms, i.e., for each hypernym, we find the sense which synset maximizes the similarity to the set of gathered hypernyms (Section~\ref{sub:hsparse}).

Additionally, we use distributional word representations to transform the sparse synset representations into dense synset representations. We obtain such representations by aggregating the word embeddings corresponding to the elements of synsets and sets of hypernyms (Section~\ref{sub:hdense}). Finally, we generate the sense-aware hyponym-hypernym pairs by computing cross products (Section~\ref{sub:linking}).

Let $V$ be a vocabulary of ambiguous words, i.e., a set of all lexical units (words) in a language. Let $\mathcal{V}$ be a set of all the senses for the words in $V$. For instance, $\sense{apple}{2} \in \mathcal{V}$ is a sense of $\textit{apple} \in V$. For simplicity, we denote $\senses(w) \subseteq \mathcal{V}$ as the set of sense identifiers for each word $w \in V$. Then, we define a synset $S \in \mathcal{S}$ as a subset of $\mathcal{V}$.

Given a vocabulary $V$, we denote the input set of \textit{is-a} relationships as $R \subset V^2$. This set is provided in the form of tuples $(w, h) \in R$. Given the nature of our data, we treat the terms \textit{hyponym} $w \in V$ and \textit{hypernym} $h \in V$ in the lexicographical meaning. These lexical units have no sense labels attached, e.g., ${R = \{(\textit{cherry}, \textit{color}), (\textit{cherry}, \textit{fruit})\}}$. Thus, given a set of synsets $\mathcal{S}$ and a relation $R \subset V^2$, our goal is to construct an asymmetrical relation $\mathcal{R} \subset \mathcal{V}^2$ that represents meaningful hypernymy relationships between word \textit{senses}.

\begin{figure}[t]
  \centering
  \includegraphics[scale=.6]{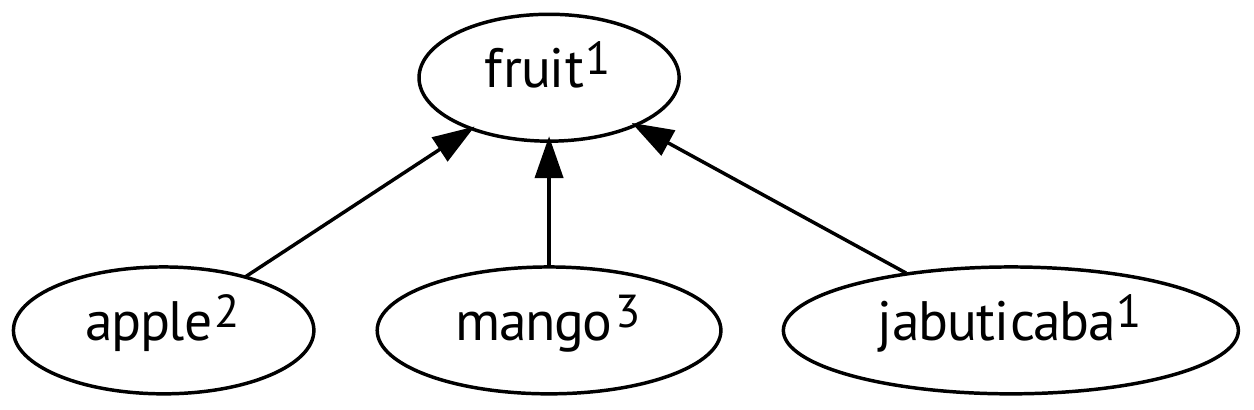}
  \caption{\label{fig:hyponyms}Disambiguated hypernymy relationships: each hypernym has a sense identifier from the pre-defined sense inventory.}
\end{figure}

The complete pseudocode for the proposed approach is presented in Algorithm~\ref{alg:linking}; the output of the algorithm is the sense-aware hypernymy relation $\mathcal{R}$ (cf.\ \figurename~\ref{fig:hyponyms}). The following sections describe various specific aspects of the approach.

\begin{figure*}
  \centering
  \includegraphics[scale=.8]{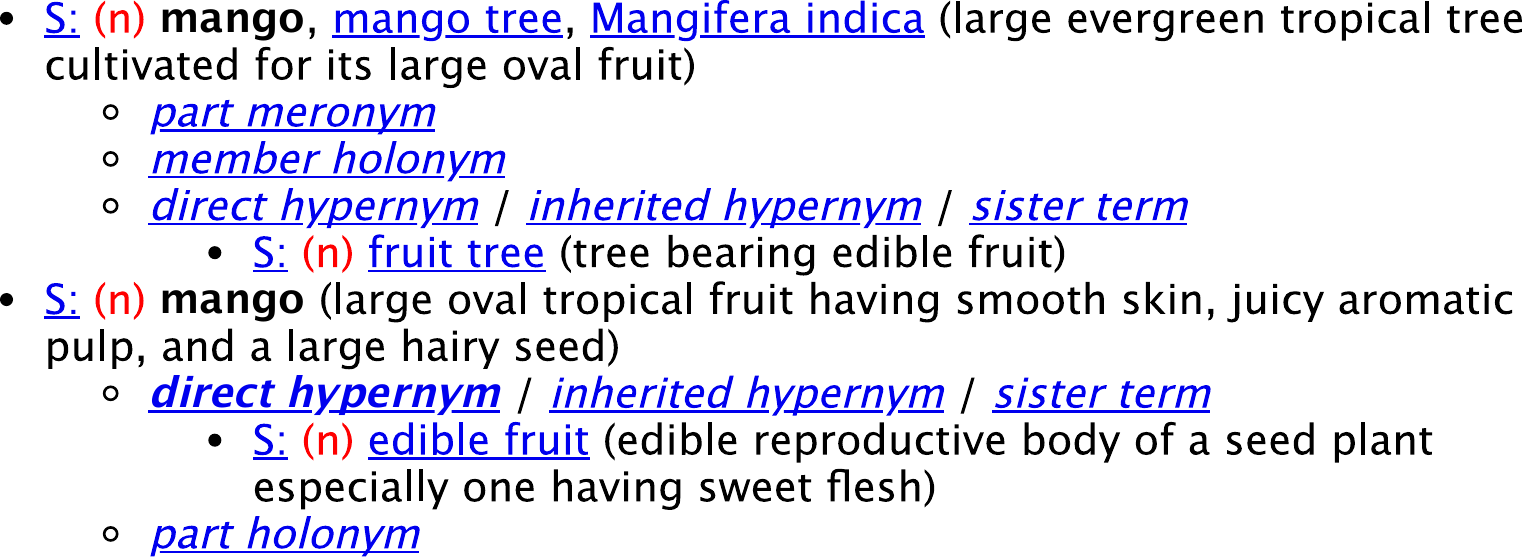}
  \caption{\label{fig:synsets} Synsets of the word ``mango'' from the Princeton WordNet and their respective hypernyms.}
\end{figure*}

\subsection{\label{sub:synsets} Obtaining Synsets}

A synset is a linguistic structure which is composed of a set of mutual synonyms, all representing the same word sense. For instance, WordNet described two following senses of the word ``mango'', which correspond to a tree and a fruit respectively, as illustrated in Figure~\ref{fig:synsets}. Note that, depending on the word sense, the word ``mango'' can have a different hypernym, which is also a synset in turn.

In our experiments, presented in this paper, we rely on synsets from the manually constructed lexical resources, such as WordNet~\cite{Fellbaum:98}, and on synsets constructed automatically from synonymy dictionaries, using the \watset{}  algorithm~\cite{Ustalov:17:acl}. 

While synonymy dictionaries can be extracted from Wiktionary and similar resources for almost any language, coverage of such dictionaries, for some languages can be still scarce. For these cases, instead of synsets, our approach can be used with distributionally induced word senses extracted from unlabelled text corpora. We explored this route in~\cite{Panchenko:2018c}. 

\subsection{\label{sub:finding}Finding Informative Synset Hypernyms}

We start with finding informative hypernyms for every synset. In real-world datasets, the input relation $R$ can contain noise in the form of mistakenly retrieved co-occurrences and various human errors. In order to get rid of these mistakes, we map every synset $S \in \mathcal{S}$ to a \textit{bag of words} $\hlabel(S) \subset V$ without sense identifiers. This synset label holds a bag of hypernyms in $R$ matching the words in $S$ as hyponyms in lines~\ref{alg:linking:label:begin}--\ref{alg:linking:label:end}:
\begin{equation}
  \hlabel(S) = \{h \in V : (w, h) \in R, w \in \words(S)\}\text{.}\!\!
  \label{eq:hsparse}
\end{equation}

In case the relation $R$ is provided with the counts of pair occurrences in a corpus, we add each occurrence into $\hlabel(S)$. Furthermore, since $\hlabel(S)$ is a bag allowing multiple occurrences of the same hypernyms for different words included to the synset, we model the variable importance of words in labels using the $\tfidf$ weighing scheme~\cite{Salton:88} in lines~\ref{alg:linking:tfidf:begin}--\ref{alg:linking:tfidf:end}:
\begin{align}
  \label{eq:tfidf}
  \tfidf(h, S, \mathcal{S}) &=
  \tf(h, S) \times \idf(h, \mathcal{S})\text{,}\\
  \label{eq:idf}
  \tf(h, S) &=
  \frac{|h' \in \hlabel(S) : h = h'|}{|\hlabel(S)|}\text{,}\\
  \label{eq:tf}
  \idf(h, \mathcal{S}) &=
  \log\frac{|\mathcal{S}|}{|S' \in \mathcal{S} : h \in \hlabel(S')|}\text{.}
\end{align}

In order to ensure that the most important hypernyms are the terms that often were identified as hypernyms for the respective synset, we limit the maximal size of $\hlabel(S)$ to a parameter $n \in \mathbb{N}$. As the result of this step, each synset is provided with a set of top-$n$ hypernyms the importance of which is measured using $\tfidf$. 

\subsection{\label{sub:hsparse}Hypernym Sense Disambiguation}

The words in the synset labels are not yet provided with sense labels, so in this step, we run a word sense disambiguation procedure that is similar to the one by~\newcite{Faralli:16}. In particular, given a synset $S \in \mathcal{S}$ and its $\hlabel(S) \subseteq V$, for each hypernym $h \in \hlabel(S)$ we aim at finding the synset $S' \in \mathcal{S}$ such that it is similar to the whole $\hlabel(S)$ containing this hypernym while it is not equal to $S$.

We perform the hypernym sense disambiguation as follows. Every synset and every label are represented as sparse vectors in a vector space model that enables computing distances between the vectors~\cite{Salton:75}. Given a synset $S \in \mathcal{S}$ and its label, for each hypernym $h \in \hlabel(S)$ we iterate over all the synsets that include $h$ as a word. We maximize the cosine similarity measure between $\hlabel(S)$ and the candidate synset $S' \in \mathcal{S}$ to find the synset $\hat{S}$ the meaning of which is the most similar to $\hlabel(S)$. The following procedure is used (line~\ref{alg:linking:sparse:wsd}):
\begin{equation}
  \hat{S} = \!\!\!\!\!\!\!\!\!\!\!\!\!\!\!\underset{S' \in \mathcal{S} : \senses(h) \cap S' \neq \emptyset}{\arg\max}\!\!\!\!\!\!\!\!\!\!\!\!\!\!\!\!\ssim(\hlabel(S), \words(S'))\text{.}
  \label{eq:dhyper}
\end{equation}

Having obtained the synset $\hat{S}$ that is closest to $\hlabel(S)$, we treat $\hat{h} = \senses(h) \cap \hat{S}$ as the desired disambiguated sense of the hypernym $h \in \hlabel(S)$. This procedure is executed for every word in the label to augment the disambiguated label (lines~\ref{alg:linking:sparse:add:begin}--\ref{alg:linking:sparse:add:end}):
\begin{equation}
  \widehat{\hlabel}(S) = \widehat{\hlabel}(S) \cup \{\hat{h} \in \mathcal{V} : h \in \hlabel(S)\}
\end{equation}

The result of the label construction step is the set of disambiguated hypernyms linked to each synset. For example, consider the hypernymy label $\{\textit{fruit}, \textit{food}, \textit{cherry}\}$ and two following synsets: $\{\sense{cherry}{1}, \sense{red fruit}{1}, \sense{fruit}{1}\}$ and $\{\sense{cherry}{2}, \sense{cerise}{1}, \sense{cherry red}{1}\}$. The disambiguation procedure will choose the first sense of the word ``fruit'' in the hypernymy label because the latter synset is more similar to the given label.

\subsection{\label{sub:hdense}Embedding Synsets and Hypernyms}

In order to overcome data sparsity by retrieving more relevant senses, we use such distributional word representations as Skip-gram~\cite{Mikolov:13}. We embed synsets and their labels in a low-dimensional vector space to perform matching. This matching makes it possible to produce more sense-aware hypernymy pairs as it captures the hierarchical relationships between synsets through their labels. Given a word $w \in V$, we denote as $\vec{w} \in \mathbb{R}^d$ a $d$-dimensional vector representation of this word.

Given the empirical evidence of the fact that a simple averaging of word embeddings yields a reasonable vector representation~\cite{Socher:13}, we follow the SenseGram approach by \newcite{Pelevina:16} to compute synset embeddings. We perform \textit{unweighted} pooling as the words constituting synsets are equally important (line~\ref{alg:linking:dense:svec}):
\begin{equation}
  \vec{S} = \frac{\sum_{w \in \words(S)} \vec{w}}{|S|}\text{.}
  \label{eq:svector}
\end{equation}

In contrast to the approach we use to embed synsets, we perform \textit{weighted} pooling of the word embeddings to compute the label embeddings. Like the weights, we use $\tfidf$ scores produced at the synset labeling stage (Section~\ref{sub:finding}). Thus, each $\hlabel(S)$ is mapped to the following low-dimensional vector (line~\ref{alg:linking:dense:lvec}):
\begin{equation}
  \overrightarrow{\hlabel}(S) = \frac{\sum_{h \in \hlabel(S)} \tfidf(h, S, \mathcal{S}) \cdot \vec{h}}{\sum_{h \in \hlabel(S)} \tfidf(h, S, \mathcal{S})}\text{.}
  \label{eq:hvector}
\end{equation}

Now, we use a \textit{top-down} procedure for establishing relationships between the synsets as follows. We represent all the synsets $\mathcal{S}$ and all their labels in \textit{the same} vector space. Then, for each synset label, we search for the $k \in \mathbb{N}$ nearest neighbors of the label vector. In case we find a synset among the top neighbors, we treat it as the set of hypernyms of the given synset. Specifically, given a synset $S \in \mathcal{S}$ and its $\overrightarrow{\hlabel}(S) \in \mathbb{R}^d$, we extract a set of nearest neighbors $\NN_k(\overrightarrow{\hlabel}(S))$. Each element of the result set can be either a synset or a label. We do not take into account the neighbors that are labels. We also exclude the input synset from the result set. Thus, for the synset $\hat{S}$ we use a disambiguation procedure shown in line~\ref{alg:linking:dense:wsd}:
\begin{equation}
  \hat{S} = \!\!\!\!\!\!\!\!\!\!\!\!\!\!\underset{S' \in \NN_k(\overrightarrow{\hlabel}(S)) \cap \mathcal{S} \setminus \{S\}}{\arg\max\ssim}\!\!\!\!\!\!\!\!\!\!\!\!\!\! (\overrightarrow{\hlabel}(S), \vec{S}')\text{.}
\end{equation}

Additionally, we require that no candidate synset includes more than $m \in \mathbb{N}$ words as it can hardly represent a reasonable set of synonyms. Finally, to each $S \in \mathcal{S}$ we assign ${\widehat{\hlabel}(S) = \hat{S}}$ in lines~\ref{alg:linking:dense:size:begin}--\ref{alg:linking:dense:size:end}. In case no synsets are found, we skip $S$.

During prototyping, we tried the bottom-up procedure of searching a label given a synset. Our experiments showed that such a procedure is inefficient and fails to provide a reasonable matching.

\begin{table*}[t]
\centering
\caption{\label{tab:hypernyms}Hypernyms used to construct labels of the input synsets, the frequency threshold for Hearst Patterns is denoted as $f \in \mathbb{N}$.}
\begin{tabular}{cp{120mm}r}\toprule
\textbf{Language} & \textbf{Name} & \textbf{\# pairs} \\\midrule
\multirow{3}{*}{\rotatebox[origin=c]{90}{English}} & Wiktionary & $62\,866$ \\
& Hearst Patterns ($f \geq 100$) & $39\,650$ \\
& \textsc{All} (Wiktionary + Hearst Patterns) & $102\,516$ \\\midrule
\multirow{4}{*}{\rotatebox[origin=c]{90}{Russian}} & Wiktionary & $185\,257$ \\
& Hearst Patterns ($f \geq 30$) & $10\,458$ \\
& Small Academic Dictionary & $38\,661$ \\
& \textsc{All} (Wiktionary + Small Academic Dictionary + Hearst Patterns) & $234\,376$ \\\bottomrule
\end{tabular}
\end{table*}

\subsection{\label{sub:linking}Generation of Hypernymy Pairs}

We generate an output set of sense-aware hyponym-hypernym pairs $\mathcal{R} \subset \mathcal{V}^2$ by computing a cross product between the set of synsets and the set the labels corresponding to them (line~\ref{alg:alinking:return}):
\begin{equation}
  \mathcal{R} = \bigcup_{S \in \mathcal{S}} S \times \widehat{\hlabel}(S)\text{.}
  \label{eq:output}
\end{equation}

As the result, the example in \figurename~\ref{fig:hyponyms} will be transformed into the following relation $\mathcal{R}$:
\begin{tabular}{*{2}{p{.45\columnwidth}}}\toprule
\textbf{Hyponym Sense} & \textbf{Hypernym Sense} \\\midrule
\sense{apple}{2}       & \sense{fruit}{1} \\
\sense{mango}{3}       & \sense{fruit}{1} \\
\sense{jabuticaba}{1}  & \sense{fruit}{1} \\\bottomrule
\end{tabular}

\section{\label{sec:evaluation}Evaluation}

We conduct two experiments based on well-known gold standards to address the following research questions:
\begin{description}
  \item[\textbf{RQ1}] How well does the proposed approach generalize the hypernyms given the synsets of the gold standard?
  \item[\textbf{RQ2}] How well does the proposed approach generalize the hypernyms given the synsets not belonging to the gold standard?
\end{description}

We run our experiments on two different languages, namely English, for which a large amount of lexical semantic resources are available, and Russian, which is an under-resourced natural language. We report the performance of two configurations of our approach. The first configuration, \textit{Sparse}, excludes the embedding approach described in Section~\ref{sub:hdense} (lines~\ref{alg:linking:dense:svec}--\ref{alg:linking:dense:size:end}). The second configuration, \textit{Full}, is a complete setup of our approach, which includes the relation extracted with the \textit{Sparse} configuration and further extends them with relations extracted using synset-hypernym embedding matching mechanism.

\subsection{Experimental Setup}

Given a gold standard taxonomy, composed of hypernymy relations, one can evaluate the quality of the automatically extracted hypernyms by comparing them to this resource. A common evaluation measure for assessing taxonomies is the cumulative Fowlkes--Mallows index proposed by \newcite{Velardi:13}. However, this measure cannot be applied for relatively large graphs like ours due to running a depth-first search (DFS) algorithm to split the input directed graph into levels. Since our graphs have hundreds of thousands of nodes (cf.\ \tablename~\ref{tab:hypernyms}), this approach is not tractable in reasonable time unlike in the evaluation by \newcite{Bordea:16} that was applied to much smaller graphs. To make our evaluation possible, we perform directed path existence checks in the graphs instead of the DFS algorithm execution. In particular, we rely on precision, recall, F-score w.r.t. a sense-aware gold standard set of hypernyms. For that, sense labels are removed from the compared methods and then an \textit{is-a} pair $(w, h) \in R$ is considered as predicted correctly \textit{if and only if there is a path from some sense of $w$ to some sense of $h$ in the gold standard dataset}. Let $G = (V_G, E_G)$ be the gold standard taxonomy and $H = (V, E)$ be the taxonomy to evaluate against $G$. Let  $u \overset{G}{\rightarrow} v$ be the directed path from the node $u$ to the node $v$ in $G$. Then, we define the numbers of positive and negative answers as follows:
\begin{alignat}{2}
  \mathrm{TP} &= |(u, v) \in E &:  \exists u \overset{G}{\rightarrow} v|\text{,}\\
  \mathrm{FP} &= |(u, v) \in E &: \nexists u \overset{G}{\rightarrow} v|\text{,}\\
  \mathrm{FN} &= |(u, v) \in E_G &: \nexists u \overset{H}{\rightarrow} v|\text{,}
\end{alignat}
where $\mathrm{TP}$ is the number of true positives, $\mathrm{FP}$ is the number of false positives, and $\mathrm{FN}$ is the number of false negatives. As the result, we use the standard definitions of precision as ${\mathrm{Pr} = \frac{\mathrm{TP}}{\mathrm{TP} + \mathrm{FP}}}$, recall as ${\mathrm{Re} = \frac{\mathrm{TP}}{\mathrm{TP} + \mathrm{FN}}}$, and F-score as ${\mathrm{F}_1 = \frac{2 \cdot \mathrm{Pr} \cdot \mathrm{Re}}{\mathrm{Pr} + \mathrm{Re}}}$.

Note that the presented approach could overestimate the number of true positives when the nodes are located far from each other in the gold standard. Only the words appearing both in the gold standard and in the comparable datasets are considered. The remaining words are excluded from the evaluation.

\subsection{Datasets}

The hypernymy datasets for both languages have been extracted from Wiktionary using the JWKTL tool by \newcite{Zesch:08}; the Wiktionary dump was obtained on June 1, 2018. As the non-gold datasets of synsets, we use the automatically discovered synsets published by~\newcite{Ustalov:17:acl} for both English and Russian.\footnote{\url{https://github.com/dustalov/watset/releases/tag/v1.0}}

\begin{table}[t]
\centering
\caption{\label{tab:embeddings}Skip-gram-based word embeddings used to construct synset embeddings.}
\resizebox{\linewidth}{!}{
\begin{tabular}{lllcr}\toprule
\textbf{Language} & \textbf{Dataset} & \textbf{Genre} & \textbf{Dim.} & \textbf{\# tokens} \\\midrule
English & Google News & news  & $300$ & $100 \times 10^9$ \\
Russian & RDT         & books & $500$ &  $13 \times 10^9$ \\\bottomrule
\end{tabular}
}
\end{table}

For \textit{English}, we combine two data sources: \textit{Wiktionary} and a hypernymy pair dataset obtained using \textit{Hearst Patterns} from a large text corpus. The corpus has 9 billion tokens compiled from the Wikipedia\footnote{\url{http://panchenko.me/data/joint/corpora/en59g/wikipedia.txt.gz}}, Gigaword~\cite{Graff:03}, and ukWaC~\cite{Ferraresi:08} corpora. The union of hypernyms from Wiktionary and Hearst patterns is denoted as \textsc{All}. As word embeddings for English, we use the Google News vectors.\footnote{\url{https://code.google.com/archive/p/word2vec/}} Finally, WordNet~\cite{Fellbaum:98} was used as the gold standard dataset in our experiments as a commonly used source of ground truth hypernyms.

For \textit{Russian}, we use a composition of three different hypernymy pair datasets summarized in \tablename~\ref{tab:hypernyms}: a dataset extracted from the \texttt{lib.rus.ec} electronic library using the \newcite{Hearst:92} patterns implemented for the Russian language in the PatternSim\footnote{\url{https://github.com/cental/patternsim}} toolkit~\cite{Panchenko:12:konvens}, a dataset extracted from the Russian \textit{Wiktionary}, and a dataset extracted from the sense definitions in the \textit{Small Academic Dictionary} (SAD) of the Russian language~\cite{Kiselev:15:isa}. We also consider the \textsc{All} dataset uniting \textit{Patterns}, \textit{Wiktionary} and \textit{Small Academic Dictionary}. As word embeddings, we use the Russian Distributional Thesaurus (RDT) vectors.\footnote{\url{https://russe.nlpub.org/downloads/}} Finally, as the gold standard, we use the RuWordNet\footnote{\url{http://ruwordnet.ru/en/}} lexical database for Russian~\cite{Loukachevitch:16}.

\subsection{Meta-Parameters of the Methods}

Parameter tuning during prototyping showed that the optimal parameters for English were $n=3$, $k=1$ and $m=15$ for WordNet, and $n=3$, $k=1$ and $m=20$ for {\watset}; for Russian the optimal values were $n=3$, $k=1$ and $m=20$ for all the cases. Table~\ref{tab:embeddings} briefly describes the word embedding datasets.

\section{Results and Discussion}

Tables~\ref{tab:english} and \ref{tab:russian} show the results for the first experiment on hypernymy extraction using for both languages. According to the experimental results for both languages on the \textit{gold standard synsets}, the \textit{Full} model outperforms the others in terms of recall and F-score. The improvements are due to gains in recall with respect to the input hypernyms (\textit{No Synsets}). This confirms that the proposed approach improves the quality of the input hypernymy pairs by correctly propagating the hypernymy relationships to previously non-covered words with the same meaning.

According to the experiments on the \textit{automatically induced synsets} by the {\watset} method from \newcite{Ustalov:17:acl}, the \textit{Full} model also yields the best results, the quality of the synset embeddings greatly depends on the quality of the corresponding synsets. While these synsets did not improve the quality of the hypernymy extraction for English, they show large gains for Russian. 

\begin{table}[t]
\centering
\caption{\label{tab:english}Performance of our methods on the WordNet gold standard using the synsets from WordNet (PWN) and automatically induced synsets (\watset) for English; the best overall results are boldfaced.}
\resizebox{\linewidth}{!}{
\begin{tabular}{clrrrr}\toprule
& \textbf{Method} &
\textbf{\# pairs} &
\textbf{Pr} &
\textbf{Re} &
\textbf{F\textsubscript{1}}\\\midrule
\multirow{2}{*}{\rotatebox[origin=c]{90}{PWN}}
& \textit{Full}(\textsc{All})   & $75\,894$ & $53.23$ & $\mathbf{39.95}$ & $\mathbf{45.27}$ \\
& \textit{Sparse}(\textsc{All}) & $61\,056$ & $56.78$ & $36.72$ & $44.60$ \\
\midrule
\multirow{2}{*}{\rotatebox[origin=c]{90}{\scriptsize\watset}}
& \textit{Full}(\textsc{All})   & $72\,686$ & $57.60$ & $18.93$ & $28.49$ \\
& \textit{Sparse}(\textsc{All}) & $40\,303$ & $62.42$ & $16.85$ & $26.53$ \\
\midrule
\multirow{3}{*}{\small\rotatebox[origin=c]{90}{No Synsets}}
& \textsc{All}                  & $98\,096$ & $64.84$ & $18.72$ & $29.05$ \\
& Hearst Patterns               & $38\,530$ & $\mathbf{67.09}$ & $16.57$ & $26.58$ \\
& Wiktionary                    & $59\,674$ & $46.78$ & $ 1.36$ & $ 2.64$ \\\bottomrule
\end{tabular}
}
\end{table}

Error analysis shows the improvements for Russian can be explained by higher quality input synsets for this language: some English synsets are implausible according to our human judgment. For both languages, our method improves both precision and recall compared to the union of the input hypernyms, \textsc{All}. Finally note that while the absolute numbers of precision and recall are somewhat low, especially for the Russian language, these performance scores are low even for resources constructed completely manually, e.g., Wiktionary and the Small Academic Dictionary in Table~\ref{tab:russian}. This is the result of a vocabulary mismatch between the gold standards and the input hypernymy datasets. Note that the numbers of pairs reported in Tables~\ref{tab:english} and~\ref{tab:russian} differ from the numbers presented in Table~\ref{tab:hypernyms} also due to a vocabulary mismatch.

\section{Conclusion}

In this study, we presented an unsupervised method for disambiguation and denoising of an input database of noisy ambiguous hypernyms using automatically induced synsets. Our experiments show a substantial performance boost on both gold standard datasets for English and Russian on a hypernymy extraction task. Especially supported by our results on Russian, we conclude that our approach, provided even with a set of automatically induced synsets, improves hypernymy extraction without explicit human input. The implementation\footnote{\url{https://github.com/dustalov/watlink}, \url{https://github.com/tudarmstadt-lt/sensegram/blob/master/hypernyms.ipynb}} of the proposed approach and the induced resources\footnote{\url{http://ltdata1.informatik.uni-hamburg.de/joint/hyperwatset/konvens}} are available online. Possible directions for future studies include using a different approach for synset embeddings~\cite{Rothe:15} and hypernym embeddings~\cite{Nickel:17}.

\section*{Acknowledgments}

We acknowledge the support of the Deutsche Forschungsgemeinschaft (DFG) foundation under the ``JOIN-T'' and ``ACQuA'' projects, and the Deutscher Akademischer Austauschdienst (DAAD). Finally, we are grateful to three anonymous reviewers for providing valuable comments. 

\begin{table}[t]
\centering
\caption{\label{tab:russian}Performance of our methods on the RuWordNet gold standard using the synsets from RuWordNet (RWN) and automatically induced synsets (\watset) for Russian; the best overall results are boldfaced.}
\resizebox{\linewidth}{!}{
\begin{tabular}{clrrrr}\toprule
& \textbf{Method} &
\textbf{\# pairs} &
\textbf{Pr} &
\textbf{Re} &
\textbf{F\textsubscript{1}}\\\midrule
\multirow{2}{*}{\rotatebox[origin=c]{90}{RWN}}
& \textit{Full}(\textsc{All})   & $297\,387$ & $37.65$ & $\mathbf{41.88}$ & $\mathbf{39.65}$ \\
& \textit{Sparse}(\textsc{All}) & $145\,114$ & $31.53$ & $22.02$ & $25.93$ \\
\midrule
\multirow{2}{*}{\rotatebox[origin=c]{90}{\scriptsize\watset}}
& \textit{Full}(\textsc{All})   & $281\,006$ & $25.75$ & $17.27$ & $20.67$ \\
& \textit{Sparse}(\textsc{All}) & $166\,937$ & $25.58$ & $13.83$ & $17.95$ \\
\midrule
\multirow{4}{*}{\rotatebox[origin=c]{90}{No Synsets}}
& \textsc{All}                  & $212\,766$ & $23.48$ & $ 9.81$ & $13.84$  \\
& SAD                           &  $36\,800$ & $24.41$ & $ 5.44$ & $ 8.90$ \\
& Wiktionary                    & $172\,999$ & $\mathbf{42.04}$ & $ 3.78$ & $ 6.94$ \\
& Hearst Patterns               &  $10\,458$ & $39.49$ & $ 0.62$ & $ 1.22$ \\\bottomrule
\end{tabular}
}
\end{table}

\bibliographystyle{konvens2018}
\bibliography{linking}

\end{document}